\PassOptionsToPackage{table,xcdraw}{xcolor}

\documentclass[sigconf]{acmart}
\AtBeginDocument{%
  }

\setcopyright{rightsretained}
\copyrightyear{2025}
\acmYear{2025}
\acmDOI{XXXXXXX.XXXXXXX}

\graphicspath{ {images/} }

\begin{document}

\title{CEQuest: Benchmarking Large Language Models for Construction Estimation}


\author{Yanzhao Wu}
\affiliation{%
  \institution{Florida International University}
  \city{Miami}
  \state{FL}
  \country{USA}}
\email{yawu@fiu.edu}

\author{Lufan Wang}
\affiliation{%
  \institution{Florida International University}
  \city{Miami}
  \state{FL}
  \country{USA}}
\email{lufwang@fiu.edu}

\author{Rui Liu}
\affiliation{%
  \institution{University of Florida}
  \city{Gainesville}
  \state{FL}
  \country{USA}}
\email{liurui@ufl.edu}


\newcommand{\datasetname}{CEQuest}

\begin{abstract}
    Large Language Models (LLMs) have demonstrated remarkable capabilities across a wide range of general-domain tasks. However, their effectiveness in specialized fields, such as construction, remains underexplored. In this paper, we introduce CEQuest, a novel benchmark dataset specifically designed to evaluate the performance of LLMs in answering construction-related questions, particularly in the areas of construction drawing interpretation and estimation. We conduct comprehensive experiments using five state-of-the-art LLMs, including Gemma 3, Phi4, LLaVA, Llama 3.3, and GPT-4.1, and evaluate their performance in terms of accuracy, execution time, and model size. Our experimental results demonstrate that current LLMs exhibit considerable room for improvement, highlighting the importance of integrating domain-specific knowledge into these models. To facilitate further research, we will open-source the proposed CEQuest dataset, aiming to foster the development of specialized large language models (LLMs) tailored to the construction domain. 
\end{abstract}

\ccsdesc[500]{Computing methodologies~Natural language processing}
\ccsdesc[500]{Applied computing~Architecture (buildings)}
\ccsdesc[500]{General and reference~Estimation}
\keywords{Large Language Model, Construction Estimation, Benchmarking}

\maketitle

\section{Introduction}
Large Language Models (LLMs) have revolutionized Artificial Intelligence (AI), demonstrating remarkable proficiency in general-domain tasks, such as natural language understanding and text summarization~\cite{liu2025exploring,kampelopoulos2025review}. However, their effectiveness in specialized fields, such as the construction domain, requires deep, domain-specific knowledge, and remains underexplored~\cite{mastering-llms-AEC}. The construction sector, with its significant economic impact~\cite{construction-stat-2025} and reliance on complex, often unstructured data such as technical drawings, presents a prime area for LLM applications~\cite{ai-construction-doc-analysis}. This paper introduces \datasetname{}, a novel benchmark dataset designed to evaluate LLM performance in construction drawing interpretation and estimation, addressing a critical need for specialized evaluation tools in this domain.

The transition from general LLMs to domain-specific models is a key frontier in AI~\cite{ling2023domain,openagi,shi2025deep}. While general LLMs excel broadly, their performance can fade in specialized areas like architecture, medicine, finance, and law, which demand nuanced understanding and precise terminology not always present in general training data~\cite{ke2025demystifying,song2025injecting,ijms111780}. Domain-specific LLMs, often fine-tuned with curated datasets, aim to overcome these limitations by providing more accurate and reliable outputs within a particular field. The development of such specialized models necessitates robust, domain-specific benchmarks to accurately measure their capabilities beyond what general benchmarks can offer~\cite{song2025injecting,llm-bench-beyond}. 
The construction industry, valued globally in trillions of dollars~\cite{construction-stat-2025}, is characterized by project-based complexities, diverse stakeholders, and vast amounts of unstructured data, such as technical drawings, schedules, specifications, and reports~\cite{blanco2023start}. Despite its scale, the sector has been slow to adopt digital technologies, leading to inefficiencies, cost overruns, and project delays~\cite{smart-construction-case-singapore,construction-slow-embrace-technology}. LLMs offer the potential to unlock value from various types of construction data, especially in interpreting complex construction drawings and improving estimation accuracy. 

\begin{figure*}[h!]
    \centering
    \includegraphics[width=0.885\textwidth]{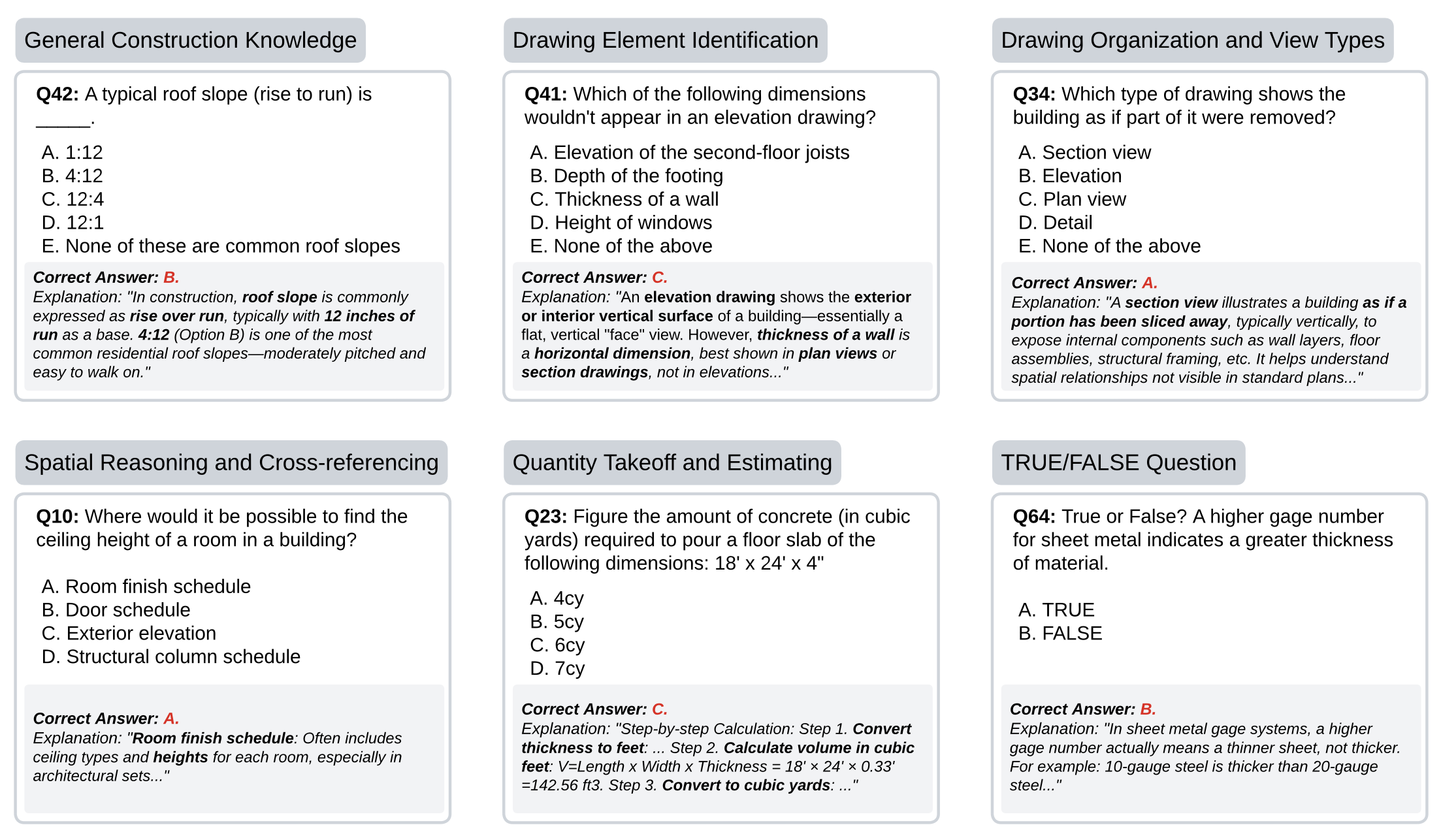}
    \caption{Sample Questions and Correct Responses.}
    \label{fig:sampleresults}
\end{figure*}

Two foundational tasks in construction that are in urgent need for AI intervention are construction drawing interpretation and estimation. Construction drawings are dense, multi-modal documents conveying spatial, structural, and material details through graphical elements and textual annotations~\cite{automation-material-takeoff-cv}. Their interpretation is challenging due to the lack of standardization, varying scales, and obscured information, requiring deep contextual understanding that often goes beyond the capabilities of traditional OCR (Optical Character Recognition) and basic computer vision techniques~\cite{ai-interpretation-construction-blueprin}. Construction estimation, involving quantity takeoff and costing, is highly dependent on accurate drawing interpretation~\cite{ghimire2024opportunities, ghimire2025framework}. Errors in either phase can lead to significant financial losses and project delays~\cite{ghimire2025framework}. While datasets like FloorPlanCAD~\cite{FloorPlanCAD} exist, they often have limitations in scale, diversity, or focus, not fully covering the breadth of information in comprehensive construction drawings needed for estimation~\cite{FloorPlanCAD}.  
On the other hand, general LLM benchmarks like MMLU~\cite{mmlu} or HellaSwag~\cite{hellaswag} are insufficient for construction-specific tasks as they do not capture the unique data modalities or reasoning required~\cite{llm-bench-beyond}. Domain-specific benchmarks like FinanceBench (finance) and LegalBench (law) have proven invaluable in their respective fields~\cite{llm-bench-beyond}, and a similar benchmark dataset is needed for construction~\cite{li2024can}.

To address this critical gap, this paper introduces \datasetname{}, a novel benchmark dataset designed to facilitate the development and evaluation of LLMs for construction drawing interpretation and estimation tasks. 
We conducted a systematic experimental analysis using \datasetname{} to benchmark five state-of-the-art (SOTA) LLMs, i.e., Gemma 3~\cite{team2025gemma}, Phi4~\cite{abdin2024phi}, LLaVA~\cite{LLaVA-2023}, Llama 3.3~\cite{grattafiori2024llama}, and GPT-4.1~\cite{gpt-4-1}. We evaluated their accuracy, execution time, and model size. 
The selection of a diverse set of models allows for a robust assessment of current LLM capabilities on these specialized construction tasks. Our key contributions are threefold. First, we developed and publicly released \datasetname{}, a novel dataset for evaluating LLMs on construction drawing interpretation and estimation question answering. Second, we provided a comprehensive empirical evaluation of five diverse LLMs on \datasetname{}, benchmarking their accuracy, execution time, and model size. Third, we provided actionable insights into current LLM capabilities and limitations in construction, highlighting the potential for specialized LLMs.

\section{Methodology}
This section describes the methodology employed in developing and evaluating the \datasetname{} dataset, including the dataset development, dataset overview, and dataset evaluation. We provide the dataset and source code at \url{https://github.com/mlsysx/CEQuest}.

\subsection{Dataset Development}

This study developed a curated dataset, \datasetname{}, which aims to cover core competencies in construction drawing interpretation and estimation, spanning foundational drawing literacy, spatial understanding, to quantity takeoff and estimating. As a pilot study, this dataset includes 164 questions in total, which were all collected and designed by domain experts, using reference materials from construction estimating textbooks, drawing interpretation guides, and real-world instructional practices in construction education. 

To facilitate continuous improvement and ensure the dataset remains relevant and comprehensive, \datasetname{} will be publicly released as an open-source project on GitHub. We will actively encourage contributions from the broader construction education and research communities. Community members are invited to submit additional questions, suggest improvements, and expand the dataset to cover emerging topics and a wider variety of question types. We aim to establish \datasetname{} as a valuable, evolving resource that supports ongoing research and educational initiatives in construction estimation. 

\subsection{Dataset Overview}

Figure~\ref{fig:sampleresults} presents a few selected sample questions in \datasetname{}. The dataset includes two types of questions: 101 (62\%) multiple-choice questions, which offer three to six labeled answer options, and 63 (38\%) TRUE/FALSE questions, which assess binary understanding of key concepts. 
The dataset is stored in a structured JSON format. Each question entry includes the following fields:
\begin{itemize} 
    \item \texttt{Problem ID (PID)}: A unique identifier for each question. 
    \item \texttt{Question}: The text of the question, clearly describing the task or query. 
    \item \texttt{Correct Answer}: The correct response, either as a boolean (TRUE/FALSE) or a labeled choice (multiple-choice). 
    \item \texttt{Options}: A list of possible answer choices (for multiple-choice questions). 
\end{itemize}
These questions are classified into five subject areas in alignment with the progressive learning stages in construction drawing interpretation and estimation. These include: 
\begin{itemize}
\item {\texttt{General Construction Knowledge}}: Focuses on broad conceptual understanding of construction processes.
\item {\texttt{Drawing Element Identification}}: Focuses on recognizing drawing elements such as scales, labels, notes, legends, materials, and building elements (e.g., walls and doors).
\item {\texttt{Drawing Organization and View Types}}: Focuses on the understanding of different drawing types (e.g., plan view, section view, detail view), projections, and drawing layouts.
\item {\texttt{Spatial Reasoning and Cross-referencing}}: Focuses on interpreting the spatial relationships of building elements within a view and connecting information across multiple drawings (e.g., matching a wall in a plan view to its section detail).
\item {\texttt{Quantity Takeoff and Estimating}}: Focuses on interpreting dimensional data to compute counts, linear, or area measurements and perform cost estimation.
\end{itemize}

\begin{figure}[t]
    \centering
    \includegraphics[width=0.5\textwidth]{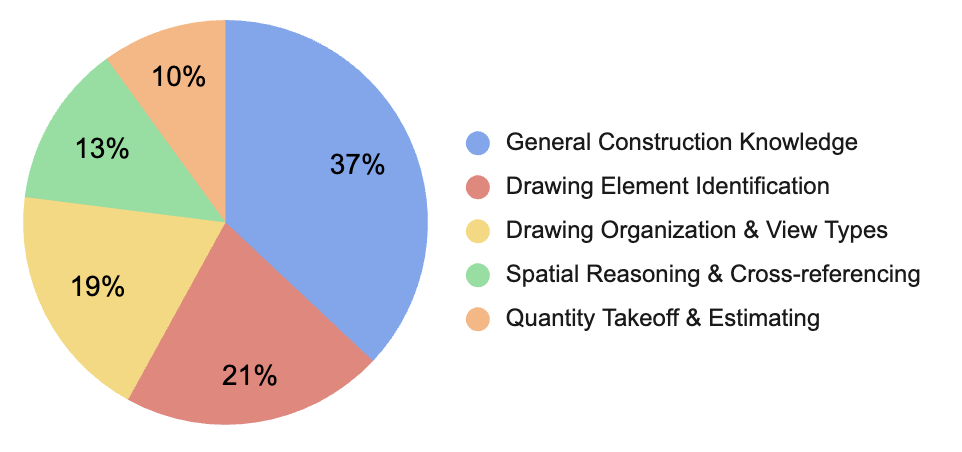}
    \caption{Distribution of Questions by Subject Area}
    \label{fig:CEQuestDist}
\end{figure}

The question distribution by subject area is demonstrated in Figure~\ref{fig:CEQuestDist}. And the statistics of the \datasetname{} dataset are summarized in Table~\ref{tab:stats}.

\subsection{Dataset Evaluation}

To facilitate reproducible evaluation, we developed a suite of evaluation functions integrated with the OpenAI API for interacting with LLMs. These functions handle querying the LLMs, parsing their outputs, and computing evaluation metrics. Specifically, we measure the following metrics: (1) \textit{Accuracy}, defined as the percentage of correctly answered questions, reflecting predictive performance; (2) \textit{Execution (Evaluation) Time}, measuring the total time taken by each LLM to complete the evaluation on \datasetname{}, indicating computational efficiency; and (3) \textit{Model Size}, referring to the storage size or number of parameters of each model, providing insights into resource requirements. 
Each LLM is evaluated on \datasetname{} using clear and consistent prompting instructions, such as \textit{``Please answer with only the letter of the correct option (A, B, C, D, E, or F)."} for multiple-choice questions. 
The benchmark framework is designed to be extensible, allowing easy integration of new LLMs, new evaluation metrics, and diverse types of construction data, thereby supporting continuous improvement and adaptation to emerging research needs in the construction domain.

\begin{table}[]
\caption{Statistics of \datasetname{} Dataset}
\label{tab:stats}
\centering
\small
\begin{tabular}{|l|c|}
\hline
\# of Questions & 164    \\ \hline
Avg. \# of Choices per Question    & 3.59 (6)  \\ \hline
Avg. \# of Sentences per Question & 1.69 (3)   \\ \hline
Avg. \# of Tokens per Question   & 14.29 (67)     \\ \hline
Avg. \# of Tokens per Choice &   1.97 (8)  \\ \hline
\end{tabular}
\begin{minipage}{0.65\linewidth}
\footnotesize
\textsuperscript{1} The numbers in the parentheses indicate the Max. 
\end{minipage}
\end{table}

\section{Experimental Analysis}

\noindent \textbf{Experimental Setup:} We evaluate five state-of-the-art LLMs on the proposed \datasetname{} dataset. The LLMs assessed include open-source LLMs, including Gemma 3 (4B)~\cite{team2025gemma}, Phi4 (14B)~\cite{abdin2024phi}, LLaVA (34B)~\cite{LLaVA-2023}, and Llama 3.3 (70B)~\cite{grattafiori2024llama}, deployed locally via Ollama~\cite{ollama}, as well as GPT-4.1~\cite{gpt-4-1} accessed via the OpenAI API. All experiments are conducted on a server equipped with an AMD Ryzen Threadripper PRO 7965WX CPU, 256GB memory, and two NVIDIA RTX 3090 GPUs. We repeated the experiments five times and reported the results as mean$\pm$std.

\begin{table}[h!]
\caption{Performance of SOTA LLMs on \datasetname{}}
\label{tab:llm-perf-comp}
\centering
\scalebox{0.97}{
\small
\begin{tabular}{|c|c|c|c|}
\hline
Model Name   & Model Size & Accuracy (\%) & Evaluation Time (s) \\ \hline
Gemma3:4b~\cite{team2025gemma}    & 3.3 GB     & 61.83\scriptsize{$\pm$0.30}  & 11.04\scriptsize{$\pm$0.73}         \\ \hline
Phi4:14b~\cite{abdin2024phi}     & 9.1 GB     & 64.02\scriptsize{$\pm$0.55}   & 70.00\scriptsize{$\pm$4.06}         \\ \hline
LLaVA:34b~\cite{LLaVA-2023}    & 20 GB      & 62.56\scriptsize{$\pm$1.06}   & 29.24\scriptsize{$\pm$4.36}         \\ \hline
Llama 3.3:70b~\cite{grattafiori2024llama} & 43 GB      & 65.37\scriptsize{$\pm$0.60}   & 175.00\scriptsize{$\pm$2.52}          \\ \hline
GPT-4.1~\cite{gpt-4-1}      & N/A\textsuperscript{1}         & \textbf{75.37}\scriptsize{$\pm$1.13}   & 86.26\scriptsize{$\pm$5.21}\textsuperscript{2}         \\ \hline
\end{tabular}
} 
\begin{minipage}{0.95\linewidth}
\footnotesize
\textsuperscript{1} The model size of GPT-4.1 is not publicly available. \\
\textsuperscript{2} The evaluation time for GPT-4.1 reflects the duration of OpenAI API calls.
\end{minipage}
\end{table}

\noindent \textbf{Performance Comparison of LLMs on \datasetname{}:}  Table~\ref{tab:llm-perf-comp} presents the performance of five state-of-the-art LLMs on our proposed \datasetname{} dataset, comparing their model size, accuracy, and evaluation time. We highlight three interesting observations. 
\textit{First,} all models achieve accuracy below 80\%, even though \datasetname{} is a relatively straightforward multi-choice dataset. This highlights substantial room for performance improvement and the pressing need to incorporate domain-specific knowledge into LLMs to enhance their predictive performance. The proposed \datasetname{} dataset thus serves as a valuable benchmark for driving further research in this area.
\textit{Second,} GPT-4.1 achieves the highest accuracy of 75.37\% among all evaluated LLMs, showing a notable performance gap between proprietary and open-source LLMs. This suggests significant opportunities to enhance open-source LLMs, particularly for addressing domain-specific challenges, such as those in the construction domain. 
\textit{Third,} model size may not always correlate with accuracy. For example, Phi-4 (14B) outperforms the larger LLaVA (34B) in terms of accuracy, demonstrating that factors such as model architecture and training data quality can be as important as model scale in determining overall LLM prediction performance. 

\begin{figure}[t]
    \centering
    \includegraphics[width=0.38\textwidth]{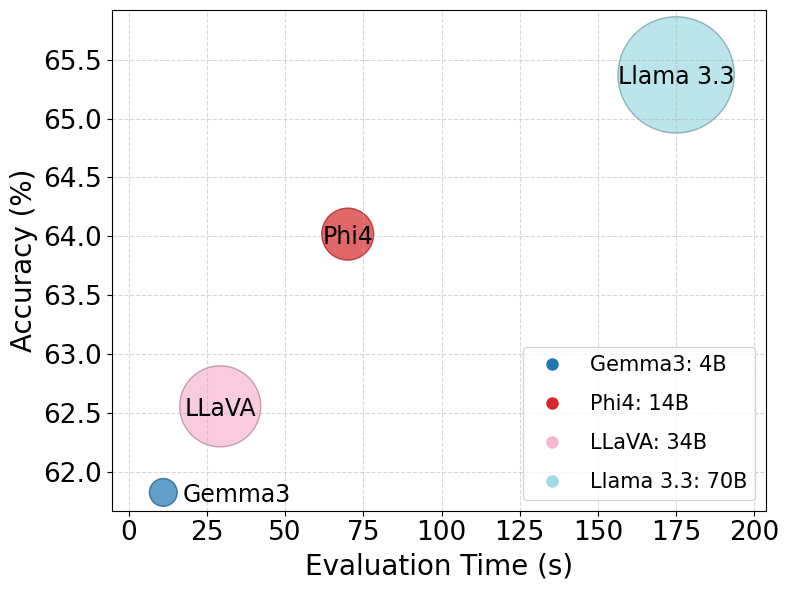}
    \vspace{-3mm}
    \caption{Comparison of Open-source LLMs on \datasetname{}.}
    \label{fig:open-source-llm-comp}
    \vspace{-3mm}
\end{figure}

Figure~\ref{fig:open-source-llm-comp} further compares four open-source LLMs evaluated on the \datasetname{} dataset. The x-axis shows the evaluation time (in seconds) while the y-axis indicates the prediction accuracy (\%). Each LLM is represented by a circle, whose size corresponds to the number of model parameters. Overall, it illustrates a clear trade-off between accuracy, evaluation time, and model size. Larger LLMs tend to achieve higher accuracy but require more computational resources and longer evaluation time.

\begin{table}[h!]
\caption{LLM Responses to Samples Questions in Figure~\ref{fig:sampleresults}}
\label{tab:llm-answers}
\centering
\scalebox{0.78}{
\small
\begin{tabular}{|c|c|c|c|c|c|c|}
\hline
PID & \footnotesize{Correct Answer} & Gemma3:4b                 & Phi4:14b                        & LLaVA:34b                 & Llama 3.3:70b             & GPT-4.1                        \\ \hline
Q10          & A              & \cellcolor[HTML]{E6EFD7}A & \cellcolor[HTML]{E6EFD7}A       & \cellcolor[HTML]{E6EFD7}A & \cellcolor[HTML]{E6EFD7}A & \cellcolor[HTML]{E6EFD7}A      \\ \hline
Q23          & C              & \cellcolor[HTML]{E6EFD7}C & \cellcolor[HTML]{EFD5CA}B       & \cellcolor[HTML]{EFD5CA}B & \cellcolor[HTML]{E6EFD7}C & \cellcolor[HTML]{EFD5CA}B \\ \hline
Q34          & A              & \cellcolor[HTML]{E6EFD7}A & \cellcolor[HTML]{E6EFD7}A       & \cellcolor[HTML]{E6EFD7}A & \cellcolor[HTML]{E6EFD7}A & \cellcolor[HTML]{E6EFD7}A      \\ \hline
Q41          & C              & \cellcolor[HTML]{E6EFD7}C & \cellcolor[HTML]{EFD5CA}B       & \cellcolor[HTML]{E6EFD7}C & \cellcolor[HTML]{EFD5CA}B & \cellcolor[HTML]{E6EFD7}C      \\ \hline
Q42          & B              & \cellcolor[HTML]{E6EFD7}B & \cellcolor[HTML]{EFD5CA}A & \cellcolor[HTML]{EFD5CA}B & \cellcolor[HTML]{E6EFD7}B & \cellcolor[HTML]{E6EFD7}B      \\ \hline
Q64          & B              & \cellcolor[HTML]{E6EFD7}B & \cellcolor[HTML]{E6EFD7}B       & \cellcolor[HTML]{E6EFD7}D & \cellcolor[HTML]{E6EFD7}B & \cellcolor[HTML]{E6EFD7}B      \\ \hline
\end{tabular}
} 
\begin{minipage}{0.95\linewidth}
\footnotesize
\textsuperscript{1} Cells highlighted in green indicate correct responses made by LLM; those in red indicate incorrect responses. 
\end{minipage}
\end{table}

\noindent \textbf{Case Study:} A few example responses generated by LLMs are presented in Table~\ref{tab:llm-answers}. The results demonstrated the capabilities of LLMs in reasoning and handling questions across various levels of complexity in construction drawing interpretation and estimation. However, our analysis also revealed several notable discrepancies and limitations.

One key challenge observed is maintaining structured responses. Although the expected output is a clear selection of the correct option (e.g., A, B, C), LLMs frequently deviated from this format. For example, the output for Q10 from LLaVA:34b was \textit{"A. Room finish schedule"}. While this answer is factually correct, it did not follow the expected format. In some cases, the models provided lengthy explanations but failed to indicate the final choice. For example, Llama 3.3 tends to provide detailed reasoning and step-by-step thought processes, but it frequently omits the actual option selection (e.g., in Q25). In other cases, explanations were included but were logically flawed or led to incorrect conclusions, despite appearing plausible -- such as Phi4's output for Q41, \textit{"An elevation drawing typically shows the exterior views of a building from all sides and includes vertical dimensions such as height and thickness. However, it does not show internal elements that are below ground level, like the depth of footings"}.

Another significant issue is the lack of domain-specific reasoning. For example, in Q23, the calculated volume of the concrete slab is 5.33 cubic yards (cy), and most LLMs (i.e., Phi4, LLaVA, and GPT-4.1) incorrectly selected Option B (5 cy), favoring the closest integer. However, in professional construction estimating practice, such quantities are typically rounded up, not to the nearest whole number, to ensure sufficient material is ordered and account for waste. This misunderstanding indicates a lack of embedded domain knowledge and practical reasoning within the model's responses, and underscores the need to enhance the domain-specific competence of LLMs for construction-domain tasks.

\section{Related Work}
We summarize related studies from two main perspectives: (1) LLM Evaluation and (2) LLM Applications in Construction. 

\noindent \textbf{LLM Evaluation:} While LLMs have demonstrated impressive performance across a variety of general-domain tasks, accurately assessing their prediction performance remains a significant challenge in AI research~\cite{liang2022holistic,chang2023survey}. 
Several recent studies have proposed general-domain benchmarks to address these limitations, including MMLU~\cite{mmlu}, HellaSwag~\cite{hellaswag}, TruthfulQA~\cite{truthfulqa}, AI2 Reasoning Challenge~\cite{arc}, and Natural Questions~\cite{natural-questions}. However, they still lack the granularity and specificity required for evaluating LLMs in highly specialized domains, such as construction~\cite{llm-bench-beyond}. 

\noindent \textbf{LLM Applications in Construction:} A growing body of research has leveraged LLMs to address challenges in construction. For example, LLMs have been applied to process unstructured construction documents (e.g., specifications, drawings, contracts, standards), enabling the extraction of critical information and automated compliance checking. These applications have been used to identify conflicts in construction documents and help reduce coordination errors during project development~\cite{llmACC}. In addition, LLMs have been employed for analyzing safety data and responding to natural language queries related to OSHA guidelines and incident records~\cite{llmSafety}. Recent studies have also explored the use of LLMs for project planning and scheduling for activity extraction, dependency mapping, and resource allocation~\cite{llmSchedule}.

Despite the growing popularity of LLM applications in construction, it still lacks benchmarks and publicly available datasets for evaluating LLM performance on critical construction tasks involving textual reasoning, multi-step spatial understanding, and interpretation of domain-specific documents~\cite{ghimire2025framework,automation-material-takeoff-cv}. Existing construction datasets, such as FloorPlanCAD~\cite{FloorPlanCAD}, primarily focus on visual recognition tasks and do not adequately address these textual reasoning and estimation capabilities. The proposed \datasetname{} dataset thus provides a novel benchmark for evaluating LLMs on construction drawing interpretation and estimation tasks. 

\section{Conclusions and Future Work}
This paper introduces \datasetname{}, a novel benchmark dataset to evaluate the performance of LLMs in construction drawing interpretation and estimation tasks. Our work addresses a critical research gap in existing benchmarks by providing a domain-specific question-answering dataset that reflects real-world challenges encountered in construction estimation. 
We conduct comprehensive experiments using five state-of-the-art LLMs, including Gemma 3, Phi4, LLaVA, Llama 3.3, and GPT-4.1, evaluating their performance in terms of accuracy, execution time, and model size. 
Our experimental results reveal a significant performance gap in state-of-the-art LLMs, underscoring the need to incorporate domain-specific knowledge to improve their effectiveness on specialized tasks. 
We will open-source the proposed \datasetname{} dataset to foster further research and development in domain-specific LLMs tailored to the construction domain. We encourage active community participation to continuously expand and refine the proposed dataset, ensuring its ongoing relevance and utility. 
In future work, we plan to expand the dataset, explore fine-tuning and retrieval-augmented generation techniques, and leverage multi-modal learning to further enhance LLM performance in the construction domain.

\bibliographystyle{ACM-Reference-Format}
\bibliography{references}

\end{document}